# Machine learning approach of Japanese composition scoring and writing aided system's design

Wanhong Huang,  DLUT
mannho_dut@yahoo.co.jp

**Abstract**

*Automatic scoring system is extremely complex for any language. Because natural language itself is a complex model. When we evaluate articles generated by natural language, we need to view the articles from many dimensions such as word features, grammatical features, semantic features, text structure and so on. Even human beings sometimes can't accurately grade a composition because different people have different opinions about the same article. But a composition scoring system can greatly assist language learners. It can make language leaner improve themselves in the process of output something.*

*Though it is still difficult for machines to directly evaluate a composition at the semantic and pragmatic levels, especially for Japanese, Chinese and other language in high context cultures, we can make machine evaluate a passage in word and grammar levels, which can as an assistance of composition rater or language learner. Especially for foreign language learners, lexical and syntactic content are usually what they are more concerned about.*

*In our experiments, we did the follows works: 1) We use word segmentation tools and dictionaries to achieve word segmentation of an article, and extract word features, as well as generate a words' complexity feature of an article. And Bow technique are used to extract the theme features. 2) We designed a Turing-complete automata model and create 300+ automatons for the grammars that appear in the JLPT examination. And extract grammars features by using these automatons. 3) We propose a statistical approach for scoring a specify theme of composition, the final score will depend on all the writings that submitted to the system. 4) We design an grammar hint function for language leaner, so that they can know currently what grammars they can use.  5) We design Writing quality analyze an statistic functions, so that the compositions' data can be used to improve Japanese Language education.*

1. Introduction

Nowadays, with the trend of globalization, more and more people begin to learn new languages. However, for many language learners, the improvement of writing ability is a difficulty. Because it is difficult for language beginner to self perceive the change of their writing ability. It is difficult to get feedback after writing. Because writing feedback often needs professional language teachers. But an automatic scoring system can help language learners get feedback to some extent. It can help language learners improve themselves. But until today, for any language, the machine is still unable to accurately grade a composition due to the complexity of natural language, and it is difficult for even human beings to accurately rating a composition at high levels of language. For example, we sometimes have difficulty evaluating a composition in semantics and pragmatics aspect. Different people may have different opinions about a composition. High context languages like Chinese and Japanese is more difficult for us to evaluating passages in high level aspect.

But fortunately, in the low-level aspects of language, such as words and grammars. The machine can effectively evaluate the quality of an article. In ***Make article a science*** [1]***,*** it proposes a way to calculate the complexity of a Japanese article by extracting the features of words in the article, and then score the article. However, this does not reflect the grammatical features of the article. We design a Turing-complete and event-driven automata module,  and then according to grammars in JLPT N5 ~ N1, about 300 automata are created to recognize the grammars. These automata are used to extract grammatical features. And with these automata, we can provide grammatical hints in the process of Japanese learners' writing. Applying machine learning algorithm to the extracted word and grammar features, we can provide a rough score evaluation for students' compositions.

1.1. Japanese-Language Proficiency Test (JLPT)

JLPT is the abbreviation of Japanese proficiency test. It is a Japanese proficiency test widely accepted in the world, which has N5~N1 totally 5 levels. And the N1 is the highest levels, which cover more than 10 thousands words' and more than 300 grammars. In our project, we create about 300 automatons for parse the corresponding grammar.

## 1.2. Kuromoji

A Japanese text segmentation library written in Java. By using it, we can segment the text conveniently and get the word's information such as pronunciation, base form, word type directly.

Figure 1. A dictionary that used by Kuromoji. We can get much information of a specific word from this dictionary. Such as word type, pronunciation, base form, category and so on. They are useful for features extraction. (For better display, this figure shows only parts of the columns. For each word, it contains more than 20 columns' property.)

In addition to lexical feature extraction, the information of these words will also play an important role in grammar analysis. Because we use automata to parse a grammar. However it can only use the appearance or the word string itself as the state transition condition of the automata. We need more information such as word type, base form and so on.

## 1.3. Dataset

The dataset we used is the composition wrote by Japanese major students in final composition examination. It's totally 281 passages and includes several different themes. All the compositions are no score label. So we need some tricks to analyze and score them.

## 2. Relative Researches

### 2.1. Calculation of article complexity by lexical features

In the *Make article a science* [1], it proposes to judge a passage's complexity by extract some features in the words. Such as total number of words, number of words in Chinese characters, number of Loanwords, the number of Japanese words, number of each word type an so on. And then impose linear regression on these regression. Even though this is a relatively simple method, it can play a certain role in the evaluation of the complexity of the article.

However, it does not introduce grammatical features. For high-level writers and beginners, we can not expect that they use the same grammar or words collocation. And foreign language beginners' composition usually contain some grammatical errors. These all contribute to a passage's quality.

### 2.2. Automatic scoring of English composition

There are many achievements and researches on English composition grading. From *The present situation and Prospect of composition automatic scoring* [11] and *Principles and comments on several automatic scoring systems for English composition* [12], there are PEG(Project Essay Grader), E-rater(Electronic Essay Rater), IEA, Criterion and IntelliMetric for English composition scoring. The scoring method of PEG is completely based on the principle of statistics. It scoring a composition by using the features like the average length of words, the total number of words, the number of commas, the number of rare words and so on. Even without considering the meaning and grammatical features of words, the correlation coefficient between the score results and the manual score result was 0.87.

The IEA is also uses statistic methods and using machine learning method of LSA. For IntelliMetric, it analyze more than 300 dimensions' vector and use non-linear model to predict the score.

On the other hand, the E-rater, which is designed by ETS. It uses general statistical methods and natural language processing technology to extract the features of compositions. It takes into account lexical diversity, text structure, theme of a passage and so on. The correlation coefficient between the scoring results and the manual scoring results was 0.97.

Even in English language, there are many achievements in automatic scoring, but the development of automatic scoring system is still challenging in agglutinating languages like Japanese and Chinese with complex grammatical structure.

We can also see that, no matter what language, it is still very difficult to grade compositions at the high semantic level. Even E-rater with high accuracy can't score a composition at the high semantic level though it takes into account the similarity of the passage theme and the required theme. However, experiment has proved that words and grammar, the characteristics of the text, has been able to make the composition scoring system to achieve a high accuracy.

### 2.3. Japanese Essay Scoring System

Currently, there is also a Japanese composition automatic scoring system called **JESS** [13], The mechanism is similar to e-rater, it uses diversity of vocabulary, percentage of big word, percentage of passive sentences and organization of the passage as the features. However, it does not make full use of the grammars and collocation of words appeared in a passage.

2.4. Bag of Words(Bow) and Word/Passage Vector

Bag of words(Bow) is an technique that can be used describe a feature of a passage. It counts occurrence of each word under a dictionary to create a long sparse vector which can represent a passage in some extent. It is like a spectrum of a passage.

However, the Bow does not consider the meaning of the words. It's better for us to train word vector or passage vector. So that we can calculate the distance between words or passages. We can also impose various algorithms on the vectors we trained.

3. Our Approach

3.1. Design of Turing-Complete Automata

For parse Japanese grammar and the collocation of words, we need design a automata model. For a Turing-Complete Automata's design, we have also described it in another paper *High-concurrency Custom-build Relational Database System's design and SQL parser design based on Turing-complete automata* [14], the basic structure is shown in Figure 2.

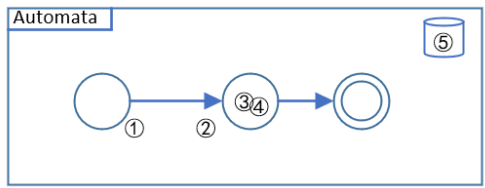

Figure 2. A simple event driven automata graph. We added a key-value storage structure (④) to the automata object. And when the state is to be shifted through an edge, the corresponding events will be triggered before(①) and after(②) the shift. You can handle the context of the automaton, or even the automaton itself.

Same to the general automata, We first need create a Automata Node class to represent Automata's Node. Then we create a Automata class. Each instance of it represent an entire automata. Different from PDA, which maintain a stack to stock context information and it can analyze Context Free Language but can't tackle with Recursively Enumerable Language. We put a hash table to the context, which likes a storage. If you put a stack object into it, the automata will become a PDA. As stack restricts the output order of data. By using a storage like structure, it can become more flexible. We can put different kind object in to the storage to satisfy different task.

The automata can becomes Turing completing because this automata is fully event driven and the node function delegation interface is put into AutomataNode class. State Changing and State Changed Event interface is also declared in Automata Edge. We reach a node, it we also trigger a Node Reached Event, in which has a default realization that call the node function delegation to do some work. You can also rewrite it to make it call the node function delegation conditionally. And you can easily write a endless loop code, which usually means it is a Turing-complete structure and it may never halt. You can also use this structure to try out many things. Even changing the automata structure in the process of state transition. Because the programming language we use is Turing complete. In theory, anything can be done. It's also easy for us to write a dead circle like programming language like Figure 3. However, as we target is natural language, in many condition, a automata that can parse CFL( Content Free Language) is enough for us. This automata can even parse RE(Enumerating recursive) language.

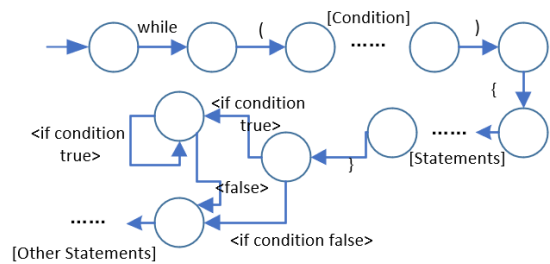

Figure 3. A automata can parse *while* statement and realize dead cycle. When receive *while,* it push a while flag to a stack in storage. Then after receive condition expression, it will create a Condition object and put it into the storage. After receive } , in Edge_BeforeShift event of next two edges, it will check the condition object and judge if the state can shift through this edge.

This automata can also be used in other various program. Such as designing a search engine of a Japanese corpus.

3.2. Using automata to recognize Japanese grammar

To evaluate the quality of a composition, grammar is a very important feature. We use about 300 important grammars in JLPT, which are also naturally graded. Some grammars only appears in high level examination.

So we need to build about 300 automata objects. For efficiency, we can use independent functions to construct some common structures, such as a substructure that parse verb-た.

Because we use strategy design pattern in the state transition of automata, we can flexibly specify the transition policy. For example, it can be transited according to a specific string, the large categories of words (such as use and body word), the general category of words (such as verbs, nouns, quantifiers), according to the prototype of words an so on.

The follow is an example that parse a grammar in N1: ~う(よう)が、~まいが A automata to parse this grammar is shown in Figure 4.

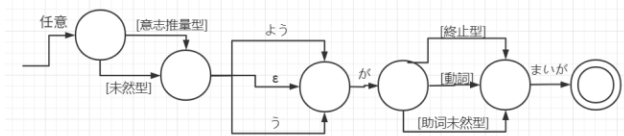

Figure 4. An automata example that parse a JLPT N1 grammar: ~う(よう)が、~まいが

Similar to the example above, we create automata object for each grammar appear in JLPT. And all the automatas are saved to a Hash Map and each automata correspond to a describe of the grammar. Parts of the codes are shown in Figure 5.

```
//10. ~（よう）が
Automata a10=builder.getNewAutomata();
builder.addWordTrans(a10,
        builder.addVolition(a10,a10.startNode),
        word: "が").setIsFinshNode(true);
automataMap.put("~う（よう）が",a10);

//11. ~（よう）が、~まいが
Automata a11=builder.getNewAutomata();
AutomataNode ab11_1=builder.addVolition(a11,a11.startNode);
AutomataNode ab11_2=builder.addWordTrans(a11, ab11_1, word: "が");
builder.addWordTrans(a11,
        builder.addAnyTrans(a11,
            ab11_2
        ), word: "まいが").setIsFinshNode(true);
//a11.matchingGrammar("彼が来ようが来まいが、パーティーは時間通りにやる。"
automataMap.put("~う（よう）が、~まいが",a11);
```

Figure 5. Some codes to create automata.

We also tried to code the automatons. And then store the coded automatons. This will greatly speed up the construction of the automata the next time the program runs. In addition, if use java language, we can also use ISerializable interface to serialize and store all automata directly.

### 3.3. Extraction of words' features

The extraction of words' feature is a relatively simple process, we just need to use some text segmentation tools like kuromoji, chasen and so on. As our project is mainly using java language, so what we choose is kuromoji.

By using text segmentation tools and dictionary, we recorded the number of verbs, nouns, conjunctions, auxiliary words, adjectives, adverbs, auxiliary verbs, total words, unique words, and Japanese words, loanwords, Chinese characters, etc.

### 3.4. Extraction of grammars' features

We have previously built nearly 300 automaton objects corresponding to nearly 300 grammars, and we have stored them in a hash table. As these grammars appear in JLPT, so they are naturally graded. We can label a grammar is belong which level in JLPT.

Then calculate the total count of each grammar, and count the total number of grammars, and count the number of grammar at each level (both total appear times and unique grammar appear times will be recorded.)

The analyze result will be put into a JSON object, which will convenient for us to do further process. The json string is as shown in Figure 6. It contains several json array represent grammar in different levels (JLPT N5~N1).

{"N1":[{"":0},{"~う（よう）が":0},{"というと":1},{"にいたる　に至る":0},{"~がてら":0},{"~た　ばかりに":0},{"をものともせずに":0},{"~くせして(くせに)":0},{"~ごとき":0},{"~こととて":0},{"~かたがた":0},{"~た　とて":0},{"~よう（う）ものなら":0},{"ではあるまいし":0},{"によって":0},{"ほかならない":0},{"ようと　まいと":0},{"にして":0},{"ぬきの　に":0},{"ゆえに　ゆえの":0},{"~としたって":0},{"~につけ":0},{"てまで":0},{"~ないではおかない":0},{"いかんにかかわらず/いかんに関わらず":0},{"~たら最後":0},{"~てたまらない":0},{"~とすれば":0},{"~ときでは":0},{"あっての":0},{"~からといって":0},{"~極まる":0},{"をもって":0},{"~すら":0},{"~といったらありはしない":0},{"を禁じえない　きんじえない":0},{"~ずにはすまない":0},{"~ずにはいられない":0},{"いかんに":0},{"~かたわら":0},{"いかんで":0},{"のみならず":0},{"まみれ":0},{"いかんだ":0},{"~ないですまない":0},{"ないものではない":0},{"~なり":0},{"~こない":0},{"~といい~といい":0},{"に至る":0},{"に至り":0},{"~たところ":0},{"なくして":0},{"にしたところで":0},{"ないまでも":0},{"にかんする　に関する":0},{"にしろ":0},{"すえに　末に":0},{"を皮切りに　をかわきりに":0},{"~かけの":0},{"いかんによらず":0},{"~ときたら":0},{"~てからというもの":0},{"を余儀なくさせる":0},{"~かけだ":0},{"に限って　に限って":0},{"ではないのだから":0},{"~ともなく":0},{"~たとたん":0},{"~はもとより":0},{"~とあれば":0},{"~ながらに":2},{"~からには":0},{"~う（よう）にも~ない":0},{"を皮切りにして":0},{"~禁じえない":0},{"のわりには":0},{"にひきかえ":0},{"~したところで　~にしたところで":0},{"にすぎない　に過ぎない":0},{"ばかりに":0},{"をよそに":0},{"てやまない":0},{"~といえども":0},{"にあたって　にあたり":0},{"~がはやいか（が早いか）":0},{"に至って":0},{"と思うと?と思ったら?とおもうと?とおもったら":0},{"めぐる":0},{"までもない":0},{"~ばなし":0},{"そくし　即し":0},{"

Figure 6. Grammars analyze result, which will be a json object.

### 3.5. Extraction of sentences' features

The average sentence length should be also counted. Because we believe that a high level language leaner can write longer sentence.

### 3.6. Extraction of theme features

Because our dataset is no score label. We can't directly use the words and grammars' features we extracted. If we have a labeled dataset, we can impose machine learning algorithm to train a model to fit the features and label. However, in the condition that we have no labeled dataset, for more precise result, we choose to also extract the theme features. This is advisable because there are more or less differences in lexical and grammatical features between different topics and different kinds of writing. For example, you can't expect letters and argumentative essays to have similar lexical and grammatical features.

Therefore, it is advisable to introduce theme features and conduct unsupervised training under the same theme.

To construct a theme feature, we introduce **Bow(Bow of word)** technique. To apply Bow technique, we need a dictionary of word. There two ways for us to do this. One is use all the words from a Japanese dictionary, by doing this, it's convenient for us to generate a Bow of different kind of passage and evaluate a passage dynamically. However, the whole dictionary is too large, and the Bow vector will be very sparse and large. The other way is only use the words that appear in a specific theme's passages. This reduces space consumption, but also at the expense of flexibility. It is difficult to evaluate new passages dynamically. But it is feasible for tasks like composition scoring in final examination. Because in the final composition examination, often the composition is not dynamically increase. In our experiment, we use the latter approach.

### 3.7. BOW Generation

After constructing a dictionary, we can create Bow for each passage which likes Figure 7. The word bag of a essay can be generated by calculating the number of times each word appears in the dictionary.

Figure 7. Generated Bag of Words (BOW).

After the word bag vector processed by machine learning algorithms, the vector of a passage can be obtained, and the theme of a passage can be reflected by the vector. Furthermore, the semantic distance between articles can also be calculated. We can believe that the theme vector of a passage under a specific theme requirement satisfies a certain high-dimensional Gaussian distribution, so we can also use statistical methods to judge the degree of digression.

### 3.8. Static Composition Evaluation

We can then grade the compositions as we have acquired the features of words, grammar and text, and we also have got the bag of words of compositions.

First we discuss the static evaluation of compositions. Static there means the unified grading of a batch of compositions under certain theme. The number of compositions will not increase dynamically.

Because there is no score labels in our dataset, we can only use the statistical technique to grade a batch of compositions. It is difficult to dynamically grade a single composition. If there are score labels in the dataset, everything will be much easier. We can even directly use general machine learning regression models to regress the score.

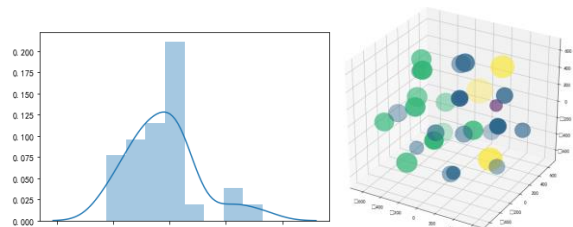

Figure 8. The features and BOW we extracted.

Figure 9. The distribution of normalized sum of each dimension (left) and the distribution of features after dimension reduce by T-SNE algorithm.

We find that after normalization, the sum of the dimensions follows a Gaussian distribution approximately (Figure 9. Left). Therefore, in the condition that the dataset without score labels, we can consider that make an evaluation based on this. Because the value of each dimension is positively correlated with the final score. The Figure 9. right graph shows that this method is feasible. The size of the dots in the right figure depends on the sum of the normalized dimensions. And the color depends on which the range of the sum the pot is on.

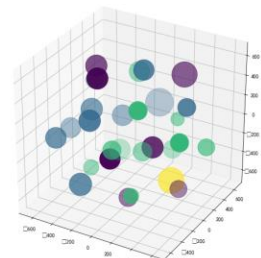

Figure 10. Clustering results using k-means on normalized features. The clustering result is also reasonable.

Figure 10. shows a clustering result by k-means algorithm. The clustering is also useful for us to grade a composition.

For grading, it's a classification problem. There two ways we can choose to use. The first way is use the normalized features' sum to decide the grade. The grade depend on which distribution range the sum in. The other way is apply clustering algorithm on features. And then for each cluster, calculate the average value and then compare to other clusters and finally attach a proper grade label likes 'A', 'B', 'C' or 'D'.

For scoring, it's a regression problem, it's hard for us to do this because no score label in the dataset. We can't use machine learning algorithms directly like common. But these also ways we can use. We have seen that the sum of the normalized feature is nearly obey a Gaussian distribution. So we can fit a gaussian distribution $N(\mu, \sigma^2)$ for it. And then, for a normalized features' sum x , calculate cumulative density function $P(X \leq x)$, which $X \sim N(\mu, \sigma^2)$. And then multiply it by a score range. For example, when the score range is expected to be between 50 and 100, the result is $50 + (100 - 50) * P(X \leq x)$.

Another way is that, we also first apply clustering algorithm to the features. Like the means that have described previously, classificate the composition to several grade. And then for each grade definite a score range. For example. A grade corresponding to 90~100, B grade corresponding to 80~90 an so on. Then for the data in each grade. We also fit a Gaussian distribution for it, that like previous method, calculating the $P(x \leq X)$ and for range $[a, b]$ , apply the formula that $a + (b - a) P(x \leq X)$.

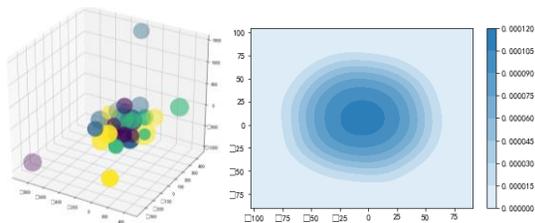

Figure 11. Text vectors that after dimension reduce. Some oblivious outliers are appear in the scatter graph (left). On the right is the 2-dimension vector data's Kernel Density Estimation.

In order to score better, we still need to consider if a passage is digress. Still, we are not know where the theme vector is. But we have reason to believe that most of the composition's vector is near the correct theme vector. So our task is to find outliers like the 4 pointers in Figure 11. To find these point, we calculate the distance between the vectors and vectors' center. If a vector is too far from the center, it will be considered as an outlier. And the final grade of the corresponding composition will go down or the score will relatively decrease.

For text vector, there are many ways for us to train it. We can use Bow and then apply a dimension reduce. We can also use Skip-Gram/CBOW to train word vector, and use some deep learning model like TextCNN, Seq2Seq, RNN. We can also use TF-IDF as weight and use weighted average of the word vectors as a text's vector.

3.9. Dynamic Composition Evaluation

For dynamic composition evaluation, the methods in 3.8 can also be used. But we need a whole dictionary for us to calculate Bow and vector of word or text. In this case, the initial vector will have more roughly 100,000 dimension, which will have a large space consumption.